\documentclass[11pt,a4paper]{article}
\usepackage[hyperref]{emnlp-ijcnlp-2019}
\usepackage{times}
\usepackage{latexsym}
\usepackage{xcolor}
\usepackage{soul}
\usepackage[colorinlistoftodos]{todonotes}
\usepackage{comment}
\usepackage{amssymb}
\usepackage{amsmath}
\usepackage{multirow}
\usepackage{footnote}
\makesavenoteenv{tabular}
\makesavenoteenv{table}
\usepackage{url}
\usepackage{graphicx}
\usepackage{mathtools}
\usepackage{float}

\aclfinalcopy 

\newcommand*\samethanks[1][\value{footnote}]{\footnotemark[#1]}

\makeatletter
\def\ps@myheadings{%
    \let\@oddfoot\@empty\let\@evenfoot\@empty
    \def\@evenhead{\thepage\hfil\slshape\leftmark}%
    \def\@oddhead{{\slshape\rightmark}\hfil\thepage}%
    \let\@mkboth\@gobbletwo
    \let\sectionmark\@gobble
    \let\subsectionmark\@gobble
    }
  \if@titlepage
  \renewcommand\maketitle{\begin{titlepage}%
  \let\footnotesize\small
  \let\footnoterule\relax
  \let \footnote \thanks
  \null\vfil
  \vskip 60\p@
  \begin{center}%
    {\LARGE \@title \par}%
    \vskip 3em%
    {\large
     \lineskip .75em%
      \begin{tabular}[t]{c}%
        \@author
      \end{tabular}\par}%
      \vskip 1.5em%
    {\large \@date \par}
  \end{center}\par
  \@thanks
  \vfil\null
  \end{titlepage}%
  \setcounter{footnote}{0}%
}
\else
\renewcommand\maketitle{\par
  \begingroup
    \renewcommand\thefootnote{\@fnsymbol\c@footnote}%
    \def\@makefnmark{\rlap{\@textsuperscript{\normalfont\@thefnmark}}}%
    \long\def\@makefntext##1{\parindent 1em\noindent
            \hb@xt@1.8em{%
                \hss\@textsuperscript{\normalfont\@thefnmark}}##1}%
    \if@twocolumn
      \ifnum \col@number=\@ne
        \@maketitle
      \else
        \twocolumn[\@maketitle]%
      \fi
    \else
      \newpage
      \global\@topnum\z@   
      \@maketitle
    \fi
    \thispagestyle{plain}\@thanks
  \endgroup
  \setcounter{footnote}{0}%
}
\makeatother

\title{Interpretable Word Embeddings via Informative Priors}

\author{Miriam Hurtado Bodell\thanks{ \hspace{2mm} Equal contribution} 
 \\
 Link\"oping University 
  \\
 {\tt \small{miriam.hurtado.bodell@liu.se}} 
 \\
 \And
 Martin Arvidsson\samethanks[1]
 \\
    Link\"oping University 
    \\
 {\tt \small{martin.arvidsson@liu.se}} 
 \\
 \And
 M{\aa}ns Magnusson
 \\
 Aalto University \\
 {\tt \small{mans.magnusson@aalto.fi}}
 \\
 }

\date{}

\begin{document}

\maketitle

\begin{abstract}
Word embeddings have demonstrated strong performance on NLP tasks. However, lack of interpretability and the unsupervised nature of word embeddings have limited their use within computational social science and digital humanities. 
We propose the use of informative priors to create interpretable and domain-informed dimensions for probabilistic word embeddings.
Experimental results show that sensible priors can capture latent semantic concepts better than or on-par with the current state of the art, while retaining the simplicity and generalizability of using priors.
\end{abstract}

\section{Introduction}

Increased availability of large digitized corpora and significant developments in natural language processing (NLP) have sparked a growing interest within computational social science and digital humanities (CSSDH) to use computational methods for textual data
\citep{laver2003extracting,grimmer2010bayesian,dimaggio2013exploiting,jockers2013significant,tsur2015frame}.
Word embeddings, a family of unsupervised methods for representing words as dense vectors \citep{mikolov2013distributed, pennington2014glove}, are one such development. Although word embeddings have demonstrated strong performance on NLP tasks \citep{DBLP:journals/corr/abs-1301-3781,mikolov2013linguistic}, they have yet to gain widespread attention within CSSDH.

We believe two key limitations can help explain the lack of applications within CSSDH.
First, since the dimensions of the word embeddings are largely uninterpretable, it is not clear how to disentangle why words are similar. 
Substantive interpretability is key for CSSDH research, and thus, the lack thereof is a major limitation.
Second, off-the-shelf word embedding models generally lack a channel through which substantive research questions can be incorporated.

To improve interpretability, previous research suggests using sparsity constraints \citep{murphy2012learning, sun2016sparse,faruqui-etal-2015-sparse} and rotation techniques \citep{park2017rotated, rothe2016word,dufter2019analytical}. 
Other work considers dimension-specific constraints to remove gender-bias, and, as a by-product, improve interpretability \cite{zhao-etal-2018-learning}. 
Within CSSDH, previous work derives interpretable dimensions via post-processing in the form of antonym-pair vector algebra \citep{kozlowski2018geometry,garg2018word} and ideal-point anchoring of antonym word-pairs \citep{lauretig-2019-identification}.
Recent formulations of word embeddings as probabilistic models \citep{DBLP:journals/corr/VilnisM14, rudolph2016exponential, barkan2017bayesian, havrylov2018embedding} enable the incorporation of domain knowledge through priors.
In this paper, we add to the literature on interpretable word embeddings, proposing a novel use of \textit{informative priors} to create predefined interpretable dimensions -- thus leveraging the expressiveness and generalizeability of the probabilistic framework.

\section{Informative Priors in Word Embeddings}
The central idea of this paper is to use informative priors to restrict the degree to which different words can inhabit different dimensions, such that one or more dimensions become interpretable and connected to one's research interest. 
Specifically, we place informative priors on word types that we expect to discriminate on a particular dimension, e.g. \textit{man-woman} for a gender dimension. 

Let $\mathcal{V}_{+}$ and $\mathcal{V}_{-}$ be the set of \textit{anchor word types} that informative priors is placed on, e.g. $  \mathcal{V}_{+} = \{man\}$ and $\mathcal{V}_{-} = \{woman\} $. 
Also, let $\mathcal{V}_{\pm} = \mathcal{V}_{+} \cup \mathcal{V}_{-}$, and let $\overline{\mathcal{V}}_{\pm}$ be the word types without any informative priors, i.e. the complement set.

Given a corpora with vocabulary size $V$, we represent each token $x_i \in \{0,1\}^{V}$ as a one-hot vector with a single nonzero entry at $v$. Here, $v \in \mathcal{V}$ represents the word type at position $i$ in the text. Following \citet{rudolph2016exponential}, we model each individual entry $x_{iv} \in \{0,1\}$ of the one-hot vectors conditional on its context $\mathbf{x}_{c_{i}}$, i.e the tokens surrounding $x_i$, where $c_{i}$ denotes the positions belonging to the context.

Each word type is associated with an \textit{embedding vector}, $\rho_{v} \in {\rm I\!R^K}$, which governs the distribution of $x_{i}$, and a \textit{context vector}, $\alpha_{v} \in {\rm I\!R^K}$, governing the distributions of the tokens for which $x_{i}$ is part of $\mathbf{x}_{c_{i}}$. 
The conditional probability of $x_{iv}$ is modelled as a linear combination of the embedding and context vectors, i.e. 
\begin{align} 
\label{eq0}
\begin{array}{rll}
p(x_{iv}|\mathbf{x}_{c_{i}}) &\sim& \text{Bernoulli}(\eta_{iv}) \,,
\end{array}
\end{align}
where
\begin{align} 
\begin{array}{rll}
\eta_{iv} &=& \text{logit}^{-1}[\rho_{v}^{\intercal} (\sum_{j\in c_{i}}\sum_{v'} \alpha_{jv'}x_{jv'})]\,,\nonumber
\end{array}
\end{align}
with logit serving as the link function. 
\citet{rudolph2016exponential} place a zero-centered Gaussian prior with variance $\sigma$ on the embedding and context vectors. 
Letting $ \theta_{v} = \{\rho_{v}, \alpha_{v}\}$, this translates to
\begin{equation} 
\label{eq1}
\theta_{v} \sim \mathcal{N}(0,\sigma)\,.
\end{equation}

In addition, since evolution of semantic concepts are of special interest in CSSDH \cite{tahmasebi2015visions}, we also consider a dynamic word embedding model to capture temporal dynamics in the dimension of interest.
We follow the specification in \citet{rudolph2017dynamic} which extends Eq. (\ref{eq0}) by associating each token with a time slice $t$, and fit separate
$\rho_{v}^{(t)} \in {\rm I\!R^K}$ for each $t$. Thus, $\theta_{v}^{(t)} = \{\rho_{v}^{(t)}, \alpha_{v}\}$. To share statistical strength between time-points, a Gaussian random walk prior is placed on $\rho_{v}^{(t)}$, i.e.
\begin{equation} 
\label{eq2}
\begin{array}{rcl}
\rho_{v}^{(t)} \sim \mathcal{N}(\rho_{v}^{(t-1)}, \sigma_{d} I)\,.
\end{array}
\end{equation}

Where $\sigma_d = \frac{\sigma}{100}$, as in  \citet{rudolph2017dynamic}, determines the smoothness of the trajectories. This shows how, in contrast to the state of the art \cite{kozlowski2018geometry,garg2018word,lauretig-2019-identification, zhao-etal-2018-learning}, informative priors allow easy integration with other, more complex, probabilistic models.

\subsection{The Standard Basis Prior}
In the following sections, we introduce a number of prior specifications that differ in how they restrict the degree to which words can occupy different dimensions.
Letting $K$ represent the dimension that we want to make interpretable, and dimensions $1:K-1$ be standard word embedding dimensions not subject to interpretation, we define our first prior specification, the \textit{Standard Basis Prior}, as
\begin{equation} 
\label{ps0}
\begin{array}{lcl}
\theta_{\overline{\mathcal{V}}_{\pm}}^{1:K} \sim  \mathcal{N}(0, \sigma) & & \theta_{\mathcal{V}_{+}}^{K} \sim  \mathcal{N}(1, \gamma) \\[6pt]
\theta_{\mathcal{V}_{\pm}}^{1:K-1} \sim  \mathcal{N}(0, \omega)& &
\theta_{\mathcal{V}_{-}}^{K} \sim  \mathcal{N}(-1, \gamma)\,,
\end{array}
\end{equation}
where $\theta_{\mathcal{V}_{+}}^{K}$ and $\theta_{\mathcal{V}_{-}}^{K}$ are priors on the dimension of interest ($K$) of $\rho_{v}$ and $\alpha_{v}$ for word types in $\mathcal{V}_{+}$ and $\mathcal{V}_{-}$ respectively, and where $\theta_{\mathcal{V}_{\pm}}^{1:K-1}$ is the shared prior for all anchor word types on dimensions $1:K-1$. Finally, $\theta_{\overline{\mathcal{V}}_{\pm}}^{1:K}$ is the standard prior from Eq. (2), placed on all dimensions for all non-anchor word types.
Hyperparameters $\omega$ and $\gamma$ are shared for $v \in \mathcal{V_{\pm}}$, controlling the strength of the prior. As $\gamma,\omega \rightarrow 0$, we force $v \in \mathcal{V}_{\pm}$ to essentially become a standard basis, defined by the word types in the prior. 
Consequently, the dot product for these vectors will be 0 for all dimensions except $K$, and thus the effect of the anchored word types on the rest of the vocabulary will exclusively depend on $K$. 
 
However, this implies that, as $\gamma,\omega \rightarrow 0$, word types within $\mathcal{V}_{-}$ and $\mathcal{V}_{+}$ obtain exactly the same word embedding. 
For example, with $\mathcal{V_{+}} = \{brother, king\}$,  \textit{brother} is treated as semantically identical to \textit{king}. To address this issue, we consider increasing $\omega$, allowing the anchor word types to exist more freely in the first $K-1$ dimensions while remaining close to $-1$ and $1$ in the $K^{th}$ dimension. This permits the use of both $brother$ and $king$ as prior anchors without assuming that they are exactly the same word. We henceforth refer to a standard basis prior with $\omega=10^{-6}$ as \textit{strict} and $\omega=1$ as \textit{weak}.

\subsection{The Truncated Prior}
A limitation of the prior specifications introduced thus far is the implicit assumptions that (i) anchor word types discriminate equally on the dimension of interest, and (ii) that we know their exact location in this dimension. 
To address this, we consider the \textit{Truncated Prior} that does not assume a basis for the anchored word types, but only that they live on the positive and negative real line as 
\begin{equation} 
\label{ps1}
\begin{array}{lcl}
\theta_{\overline{\mathcal{V}}_{\pm}}^{1:K} \sim  \mathcal{N}(0, \sigma) & &
\theta_{\mathcal{V}_{+}}^{K}  \sim  \mathcal{N}^+(0, \gamma) \\[6pt]
\theta_{\mathcal{V}_{\pm}}^{1:K-1} \sim \mathcal{N}(0, \omega) & &
\theta_{\mathcal{V}_{-}}^{K}  \sim   \mathcal{N}^-(0, \gamma)\,.
\end{array}
\end{equation}
Where $\mathcal{N}^+(0, 1)$ and $\mathcal{N}^-(0, 1)$ is the positive and negative truncated normal distribution, truncated at 0, respectively.

\subsection{Using Neutral Words}
So far, we have only considered word types known to discriminate on the dimension of interest as prior anchors. However, domain knowledge might also inform us about \textit{neutral} words in this dimension (e.g. stop words).
We thus consider placing informative priors on a third set of prior anchors $\mathcal{V}_{*}$ containing neutral word types as
\begin{equation} 
\label{ps3}
\begin{array}{rcl}
\theta_{\mathcal{V_{*}}}^{1:K-1} & \sim &  \mathcal{N}(0, \sigma)\\
\theta_{\mathcal{V_{*}}}^{K} & \sim &  \mathcal{N}(0, \psi)\,,
\end{array}
\end{equation}
where $\psi$ is the variance for dimension $K$. By guiding neutral words close to 0 in the dimension of interest, explanatory power that otherwise might have been attributed to word types in $\mathcal{V}_{*}$ will instead be allocated to other words in $\mathbf{x}_{c_{i}}$.

\section{Experiments}
Our main empirical concern is how well the proposed priors can capture meaningful latent dimensions. We summarize our empirical questions as follows:
\begin{enumerate}
    \item Which prior specification can best capture predefined dimensions?
    \item How does the best prior specification compare with the state of the art in CSSDH \cite{garg2018word,kozlowski2018geometry} (henceforth referred to as SOTA)?
\end{enumerate}

We consider two semantic dimensions, \textit{gender}, which is explored in SOTA, and \textit{sentiment}, a dimension proven difficult to capture in standard word embedding models \citep{tang2014learning}. We follow SOTA when choosing prior anchors for gender, while using the AFINN dictionary \cite{IMM2011-06010} to find prior anchors for sentiment.
To evaluate the effect of ``few'' vs. ``many'' prior anchors, we run experiments using between 2 and 276 words depending on the dimension of interest and the dataset at hand. All prior word types used in the experiments can be found in Sec. B in the supplementary material.
We follow \citet{rudolph2016exponential}, obtaining maximum a posteriori estimates of the parameters using TensorFlow \citep{abadi2015tensorflow}
with the Adam optimizer \citep{DBLP:journals/corr/KingmaB14} and negative sampling \footnote{Code is available at: https://github.com/martin-arvidsson/InterpretableWordEmbeddings}. 
We set the size of the embeddings $K=100$,
use a context window size of 8 and $\sigma=1$ throughout all experiments.

We examine the proposed priors using three commonly sized English corpora for textual analysis within CSSDH: the top 100 list of books in \citet{gutenberg}, a sample from Twitter \cite{go2009twitter} and the U.S. Congress Speeches 1981-2016 \cite{Senate}. The number of tokens ranges from 1.8M to 40M after pre-processing (see Sec. A in the supplementary material for details). The various origins, sizes and contents of these datasets work as a check of the effect of the priors in different types of corpora.

To measure the extent to which the inferred dimension reflects the semantic concept of interest, we consider how well-placed a number of pre-specified hold-out word types (not a part of $\mathcal{V}_\pm$) are in this particular dimension. Specifically, accuracy is computed as the fraction of hold-out words that are placed on the correct side of 0 on the dimension of interest in the embedding and context vectors. 
For the gender dimension, hold-out word types include the 200 most common names of the last century \cite{socsec_names} and gendered words, while for sentiment, a sample of the AFINN dictionary is used (see Sec. C in the supplementary material). 
The number of hold-out word types ranges between 213 and 275, since not all exist in each corpus. 

The experimental configurations are compared to SOTA, which we implement by (i) fitting the probabilistic word embedding model in Eq. (\ref{eq0}) and (\ref{eq1}) without informative priors, and (ii) deriving the interpretable dimension post-hoc by subtracting normalized embedding vectors of antonym-pairs, e.g. $\rho_{gender}=(\rho_{man}-\rho_{woman})+(\rho_{he}-\rho_{she})$. Comparing the sign of the cosine similarity between hold-out words and the created vector allows us to contrast the accuracy of our method with that of SOTA. To get a measure of uncertainty we calculate binomial confidence intervals using the normal approximation.

\section{Results}

\begin{figure}
    \centering
    \includegraphics[width=\columnwidth]{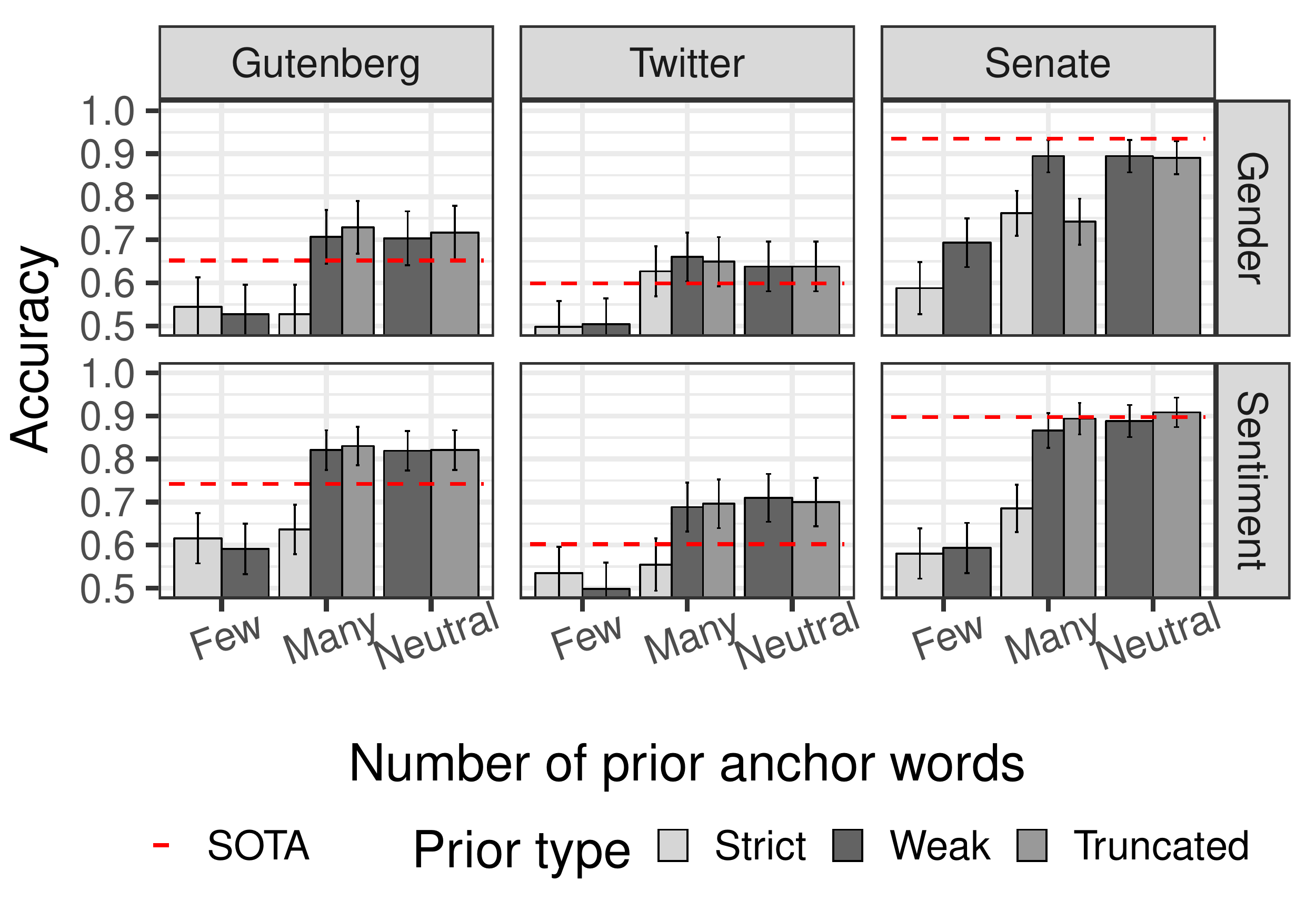}
    \caption{Comparison of prior specifications per dataset and dimension}
    \label{fig:strict_weak}
\end{figure}

We begin by comparing the strict and weak \textit{Standard Basis Priors}, with $\gamma=10^{-6}$, varying the number of anchor words used to inform the dimension of interest. 
The general pattern in Fig. \ref{fig:strict_weak} shows how (i) increasing the number of anchor words improves the accuracy, and that (ii) this improvement is much greater for the weak than for the strict \textit{Standard Basis Prior}. We explain this difference by the nature of the strict prior; it forces all anchor words to have exactly the same meaning -- which clearly is untrue. These results speak against methods that transform the vector space based on strict standard basis vectors, as in \citet{lauretig-2019-identification}.
Further noticeable is that the average accuracy is greater for the Senate corpus, which, using subsampling, we find is driven by corpora size.

Relaxing the location assumptions with the \textit{Truncated Prior} ($\gamma = 1000$) and placing informative priors on neutral words ($\psi = 0.01$) show varying degrees of improvements. 
However, due to the limited number of hold-out test words, large improvements are required for significant differences -- which only is observed for the sentiment dimension in the Senate corpus.

Fig. \hspace{-1.5mm} \ref{fig:strict_weak} further shows how our proposed approach generally performs better or on-par with SOTA. The one exception is the gender dimension in the Senate corpus. Follow-up analysis show that this difference is driven by misclassification of a cluster of ambiguous names, e.g. \textit{Madison} (founding father) and \textit{Jordan} (country). These names become correctly classified when using the full vectors, suggesting that gender has not been completely isolated in the Senate corpus (see \citet{gonen-goldberg-2019-lipstick-pig} for an in-depth discussion on issues of concept-isolation in word embeddings).

\subsection{Case Study of Semantic Change}
Using the \textit{Truncated Prior} with neutral words, we leverage the dynamic embedding model described in Eq. (\ref{eq2}) to explore temporal patterns in the Senate corpus. We set $\sigma_{d}=0.05$ to allow for the identification of abrupt temporal changes.

\begin{figure}[H]
    \centering
    \includegraphics[width=\columnwidth]{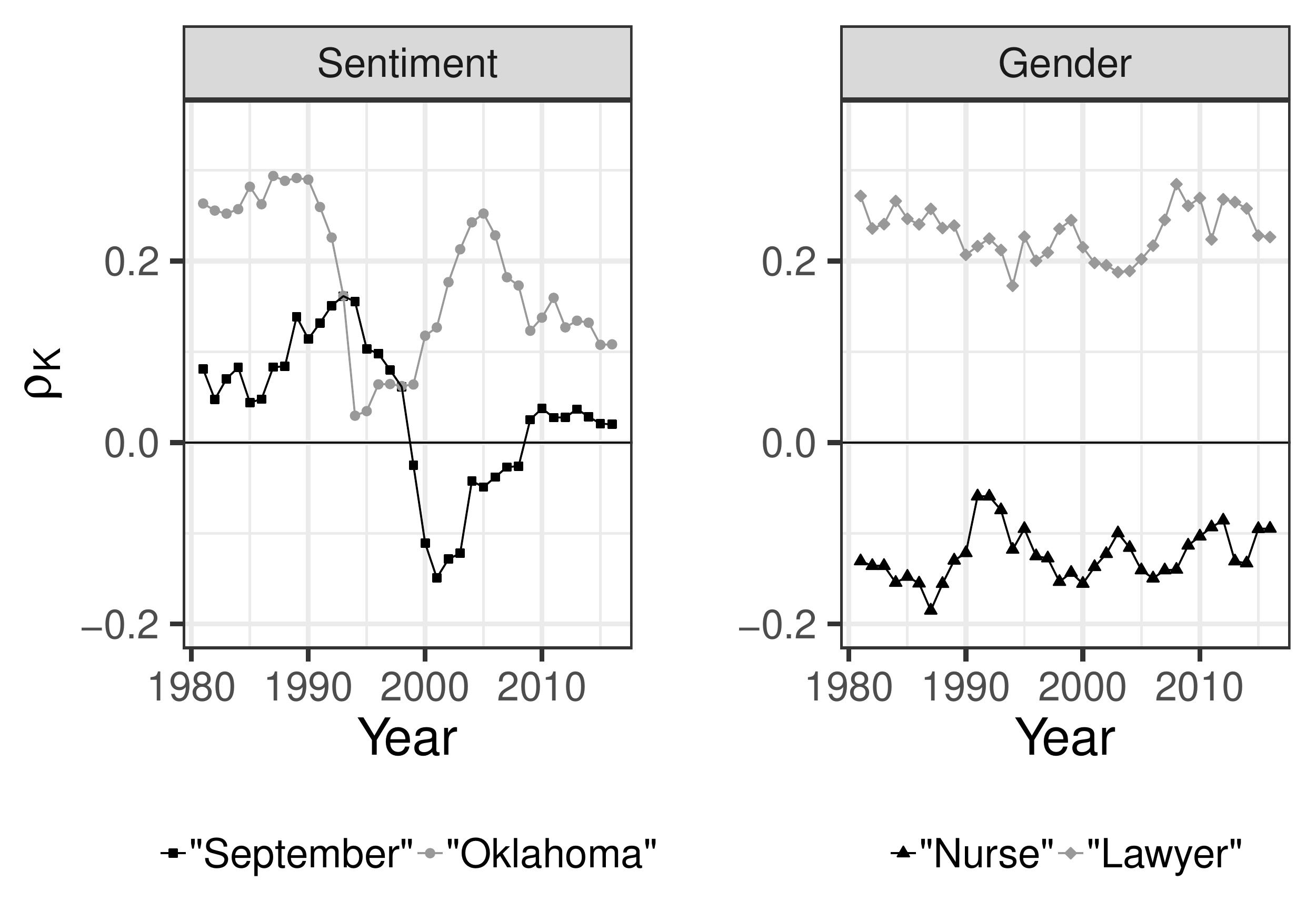}
    \caption{Gender and sentiment trajectories in the Senate corpus.}
    \label{fig:qual}
\end{figure}

The left panel of Fig. \ref{fig:qual} displays sentiment-trajectories for two words experiencing drastic three-year consecutive changes in the sentiment dimension; \textit{September} between 1999 and 2001 and \textit{Oklahoma} between 1993 and 1995, capturing two terror events: the attacks on the World Trade Center on September 11th 2001, and the Oklahoma City bombing in 1995. The change precedes the event due to the smoothness of the prior.
In the years that follow, the words gradually regain their previous sentiment -- reflecting a decline in their association with terror.

Second, we test and find support for the gender-occupation results in \citet{kozlowski2018geometry}, i.e. that occupations' temporal positioning in the gender dimension correlates well with the proportion of men and women within those fields. The right panel of Fig. \ref{fig:qual} displays the gender-dimension trajectories for the words \textit{Nurse} and \textit{Laywer}, occupations with high shares of women and men, respectively \cite{kozlowski2018geometry}.

In sum, these examples showcase how interpretability can be gained using informative priors. By isolating prespecified concepts, e.g. sentiment, into one dimension ($K$), one can infer word type-concept associations such as \textit{September-sentiment} -- for all non-anchor word types -- allowing for more meaningful within and between word type comparisons.

\section{Conclusion}
In this paper, we show how informative priors provide a simple and straightforward alternative for constructing interpretable dimensions for probabilistic word embeddings -- allowing CSSDH researchers to explore how words relate to each other in prespecified dimensions of interest, e.g. gender and sentiment.

Our results demonstrate that the biggest gains in interpretability are obtained by (i) using many prior words and (ii) increasing the degree to which they can live outside the predefined dimension. The weak \textit{Standard Basis Prior} and the \textit{Truncated Prior} overall capture the predefined dimensions with similar accuracy. 
Aligned with previous research, the experimental results indicate some issues with incomplete isolation of concepts which we believe could be addressed in future work by placing informative priors on multiple dimension to capture more complex concepts, and move away from simplistic antonym-driven dimension definitions.

Finally, while being flexible and easily extended to other probabilistic word embedding models, our approach performs better or on par with SOTA.

\section*{Acknowledgement}
This work was supported by the Swedish Research Council, grant numbers: 2018--05170 and 2018--06063, and Academy of Finland grant number 313122. We thank Marc Keuschnigg, Peter Hedstr\"om, Jacob Habinek and all reviewers for their valuable comments.

\newpage

\bibliographystyle{acl_natbib}
\bibliography{emnlp-ijcnlp-2019}

\clearpage

\title{Supplementary Material: Interpretable Word Embeddings via Informative Priors}
\author{Miriam Hurtado Bodell\samethanks[1] 
 \\
 Link\"oping University 
  \\
 {\tt \small{miriam.hurtado.bodell@liu.se}} 
 \\
 \And
 Martin Arvidsson\samethanks[1]
 \\
    Link\"oping University 
    \\
 {\tt \small{martin.arvidsson@liu.se}} 
 \\
 \And
 M{\aa}ns Magnusson
 \\
 Aalto University \\
 {\tt \small{mans.magnusson@aalto.fi}}
 \\
 }  
\maketitle

\appendix

\label{sec:supplemental}

\section{Pre-processing}

We pre-process each corpus with the following steps:

\begin{enumerate}
    \item Transform the text to lowercase, remove all punctuation, replace numbers with \textit{X} and apply stemming.
    \item Set vocabulary as the $V$ most frequent word types in the corpora, and remove other word types from the text.
    \item To speed up estimation, we follow \citep{mikolov2013distributed, rudolph2016exponential} and remove each token with probability $1-\sqrt{\frac{10^{-5}}{f_i}}$ where $f_i$ is the frequency of token $i$'s word type.
\end{enumerate}

\vspace{0.5mm}

\section{Prior anchor word types}
In this section, we list the word types that are used to specify the priors. Bold font marks the words used in the ``few''-setting, while all words are used in the ``many''-setting. Note that all words are pre-processed by stemming (see Sec. A).
\subsection{Gender}

\textbf{Positive}: \textit{\textbf{man}, men, male, father, he, him, son, boy, himself, brother, uncle, nephew}\\\\
\textbf{Negative}: \textit{\textbf{woman}, women, female, mother, she, her, daughter, girl, herself, sister, aunt, niece}

\subsection{Sentiment}
\textbf{Positive}: \textit{\textbf{posit, good}, accomplish, admir, advantag, adventur, amus, approv, ardent, attract, bargain, bliss, celebr, cherish, clean, comfort, courag, dare, defend, delight, desir, eager, ecstat, enchant, energet, enlighten, enterpris, entertain, ethic, excit, fearless, festiv, fond, freedom, gain, gallant, glori, gracious, guarante, hardi, help, hero, heroic, honest, honor, hope, humor, import, impress, improv, influenti, inspir, intellig, interest, kudo, luck, merci, merri, miracl, nobl, passion, perfect, picturesqu, play, pleas, power, prais, progress, promis, protect, reassur, recommend, rejoic, safe, satisfi, smile, solid, stabl, support, sweet, tender, thank, triumph, triumphant, unbias, visionari, willing, winner, worthi}\\\\
\textbf{Negative}: \textit{\textbf{negat, bad}, abus, accident, ach, afraid, aggrav, alien, anger, anguish, animos, annoy, antagonist, anxieti, anxious, appal, arrog, attack, aw, bastard, bias, bitch, bitter, bizarr, bomb, bore, broken, cancer, casualti, catastroph, chao, childish, clash, complain, condemn, confus, contagi, contempt, controversi, coward, cramp, crash, crime, crisi, critic, cruel, cri, damag, deadlock, death, deceiv, defect, despair, destruct, devast, die, dirt, dirti, disast, disastr, discord, dishonest, disorgan, disparag, disrupt, distract, distress, dizzi, doom, doubt, dubious, dumb, embarrass, enemi, enslav, erron, error, exagger, excus, exhaust, falsifi, farc, fear, fiasco, foolish, fraudul, frenzi, furious, haunt, helpless, hindranc, horribl, hostil, humili, hurt, hypocrit, hysteria, hyster, idiot, illeg, ill, impati, inact, inadequ, incompet, indecis, indiffer, indign, inferior, insignific, insult, irrat, lack, lag, loath, loss, lost, lurk, mad, manipul, mediocr, melancholi, menac, mischief, miseri, mistak, mistaken, mourn, murder, nasti, needi, nervous, noisi, obliter, obscen, pain, panic, passiv, pathet, pollut, powerless, prick, problem, prosecut, punish, rape, rash, reject, remors, reveng, risk, scare, scream, shaki, shit, shock, shortag, sick, sin, spite, strike, suck, suicid, suspect, suspici, terribl, terror, threat, tortur, traumat, treason, unaccept, unbeliev, uncertain, undecid, undermin, uneasi, unequ, unhappi, unjust, unsettl, unsupport, upset, urgent, weird, whore, worsen, worthless, wreck, wrong}

\subsection{Neutral}
\textbf{Neutral}: \textit{the, it, a, an, and, as, of, at, by}

\section{Hold-out test words}

In this section, we list all the word types used as hold-out words to test the accuracy of the different prior specifications.

\subsection{Gender}
\textbf{Positive}: \textit{aaron, adam, alan, albert, alexand, andrew, anthoni, arthur, benjamin, bobbi, brandon, brian, carl, charl, christoph, daniel, david, denni, dougla, edward, ethan, eugen, gabriel, georg, gerald, gregori, harold, henri, jack, jacob, jame, jason, jeremi, jerri, jess, joe, john, johnni, jonathan, jordan, jose, joseph, joshua, juan, justin, keith, kenneth, kevin, larri, lawrenc, loui, matthew, michael, nathan, nichola, noah, patrick, paul, peter, philip, randi, raymond, richard, robert, roger, russel, samuel, scott, stephen, steven, terri, thoma, timothi, vincent, walter, willi, william, zachari, king, mr, sir, princ, gentleman, gentlemen, knight, lad, mankind, monk, pope, grandfath, papa, baron, clergyman, workmen, waiter, workman, brotherhood, schoolboy, masculin, brotherinlaw, grandson, fatherinlaw, boyhood, superman, grandpapa, godfath, dad, stepfath, grandpa, greatgrandfath, cowboy, daddi, fatherhood, grandnephew, granddad, businessman, businessmen, bradley, bruce, bryan, donald, dylan, eric, gari, jeffrey, kyle, logan, ralph, ronald, roy, ryan, sean, tyler, wayn, boyfriend, batman, fanboy, boyz, playboy, stepdad, homeboy, frat, exboyfriend, boyband, babyboy, penis, granda, congressman, vicechairman}\\\\
\textbf{Negative}: \textit{abigail, amber, ami, andrea, angela, ann, anna, barbara, betti, brittani, carol, catherin, christin, cynthia, deborah, dian, diana, donna, dori, dorothi, elizabeth, emma, evelyn, gloria, hannah, heather, helen, jacquelin, jane, janet, jessica, joan, joyc, judith, juli, julia, katherin, kathleen, kelli, laura, lori, margaret, mari, maria, martha, nanci, natali, olivia, pamela, rachel, rebecca, ruth, sara, sarah, sharon, shirley, sophia, susan, teresa, theresa, victoria, mrs, ladi, queen, princess, breast, mistress, duchess, goddess, grandmoth, hostess, nun, landladi, feminin, gentlewoman, sisterinlaw, mama, stepmoth, womanhood, actress, granddaught, motherinlaw, frenchwoman, godmoth, nunneri, schoolgirl, princesss, grandmama, womankind, sisterhood, grandmamma, waitress, grandma, motherhood, greatgrandmoth, alexi, amanda, ashley, bever, brenda, carolyn, cheryl, christina, daniell, debra, denis, janic, jennif, judi, karen, kathryn, kayla, kimber, lauren, linda, lisa, madison, marilyn, megan, melissa, michell, nicol, patricia, samantha, sandra, stephani, mommi, girlfriend, momma, girli, lesbian, fangirl, babygirl, homegirl, stepmom, cowgirl, girlz, uterus, superwoman, breastfeed, feminist, grandmom, femin, congresswoman, chairwoman, servicewomen, churchwomen, businesswomen, businesswoman, vagin, femalehead, spokeswoman}

\subsection{Sentiment}
\textbf{Positive}: \textit{masterpiec, heaven, tranquil, heartfelt, clever, commend, pardon, earnest, remark, bless, treasur, pretti, faith, privileg, benefit, fortun, pleasant, encourag, loyalti, loyal, effect, ador, creativ, hug, friend, raptur, glee, joy, fame, thought, affect, fair, peac, optim, happi, grate, superior, sparkl, swift, award, eas, excel, lucki, vigil, reviv, favorit, wonder, robust, rest, innov, cheer, calm, outstand, eleg, generous, glad, hail, virtuous, confid, fascin, fun, agreeabl, beauti, hilari, comprehens, brilliant, steadfast, grace, advanc, joyous, superb, reward, respons, cute, compassion, worth, enjoy, sincer, marvel, prosper, charm, healthi, proud, top, cool, splendid}\\\\
\textbf{Negative}: \textit{defiant, retard, bankrupt, danger, propaganda, contenti, greedi, dull, insan, moodi, hardship, disregard, curs, greed, undesir, bulli, steal, misunderstand, deni, guilt, delay, vagu, broke, decept, interrupt, inhibit, lazi, liar, useless, ruin, naiv, unhealthi, outcri, shame, cruelti, weak, dire, limit, horrifi, mock, grief, choke, dread, asham, jealous, punit, angri, dark, mislead, dismal, oppress, accus, threaten, fool, fall, guilti, stab, mess, prison, bloodi, resign, distrust, stolen, worri, lonesom, fraud, horrif, scorn, dump, collaps, scandal, frantic, sabotag, skeptic, disdain, regret, stupid, hopeless, frustrat, chaotic, question, pretend, peril, disgrac, reckless, betray, blame, worn, apathi, disrespect, fatigu, miser, lunat, injustic, blind, stereotyp, thwart, unstabl, ugli, disput, ignor, weari, tire, penalti, disgust, intimid, ridicul, inabl, havoc, resent, recess, injuri, insecur, irrit, boycott, rage, persecut, cheat, disturb, sad, outrag, fright, neglect, denounc, sorrow, poverti, hell, banish, tragic, infring, silli, struggl, tragedi, selfish, nonsens, sore, restrict, uncomfort, numb, crimin, subvers, aggress, chagrin, refus, bother, wast, desper, isol, alarm, hoax, evil, poison, wick, weep, obstacl, wors, kill, lose, suffer, exclus, gross, harm, scold, screw, lone, leak, distort, depress, apprehens, meaningless, emerg, violent, fake, crush, damn, offend, disappoint, displeas, conflict, ineffect, crazi, debt, degrad, deceit, vicious, disord, timid, jeopardi, expel}

\end{document}